\documentclass[runningheads]{llncs}

\usepackage{eccv}

\usepackage{eccvabbrv}

\usepackage{graphicx}
\usepackage{booktabs}
\usepackage{algorithm}
\usepackage{algpseudocode}

\usepackage[accsupp]{axessibility}  % Improves PDF readability for those with disabilities.

\usepackage{hyperref}

\usepackage{orcidlink}

\begin{document}

\title{Heterogeneous Vision-Language Ensemble with Disagreement-Aware Reranking for Text-Based Person Anomaly Retrieval} 

\titlerunning{AIC 2026 - Track 4}

\author{Huu-An Vu\inst{1,7}\textsuperscript{*}\orcidlink{0009-0005-7580-6839} \and
Cam Tu Tran Thi\inst{2,3,7}\textsuperscript{*}\orcidlink{0009-0005-4960-6825} \and
Thanh Toan Le Ngo\inst{2,3,7}\textsuperscript{*}\orcidlink{0009-0007-4434-6942} \and
Hoang Vo\inst{3,4,7}\textsuperscript{*}\orcidlink{0009-0003-3862-9250} \and
Do Trung Hieu\inst{5,7}\textsuperscript{*}\orcidlink{0009-0001-3984-6850} \and
Hieu Dinh Trung 
Pham\inst{6}\orcidlink{0009-0007-2152-4566} \and
Khang Minh 
Le \inst{6} \and
Huy Minh Nhat 
Nguyen\inst{6}
}

% % TODO FINAL: Replace with an abbreviated list of authors.
\authorrunning{et al.}

% % TODO FINAL: Replace with your institution list.
\institute{Hanoi University of Science and Technology, Hanoi, Vietnam\\
\email{an.vh225467@sis.hust.edu.vn}\\
 \and
University of Information Technology, VNU-HCM, Ho Chi Minh City, Vietnam\\
 \and
Vietnam National University, Ho Chi Minh City, Vietnam\\
\email{\{23521704, 23521603\}@gm.uit.edu.vn}\\
\and
Ho Chi Minh city University of Science, Vietnam\\
\email{23280060@student.hcmus.edu.vn}\\
\and
VinUniversity, Hanoi, Vietnam\\
\and
Vietnamese-German University, Ho Chi Minh City, Vietnam
\and
GenAI4E Lab}

\maketitle
\begingroup
\renewcommand\thefootnote{}\footnotetext{\textsuperscript{*}~Equal contribution.}
\addtocounter{footnote}{-1}
\endgroup

\begin{abstract}

Text-based person anomaly retrieval aims to retrieve pedestrians exhibiting anomalous behaviors from a large image gallery using natural language descriptions. Compared with conventional text-based person retrieval, this task requires fine-grained reasoning over pedestrian appearance, behaviors, object interactions, and scene context, making robust cross-modal matching significantly more challenging. This paper presents the \textbf{GENAI4E} team's solution to AI City Challenge 2026 Track~4. Our framework builds upon a strong retrieval backbone and progressively integrates heterogeneous vision-language embedding models through score alignment and iterative ensemble learning, followed by disagreement-aware VLM reranking for ambiguous queries. On the official Pedestrian Anomaly Behavior (PAB) benchmark, our approach achieves \textbf{90.92\%} mAP, \textbf{85.13\%} Recall@1, \textbf{97.72\%} Recall@5, and \textbf{98.68\%} Recall@10, demonstrating the effectiveness of combining complementary vision-language representations with selective multimodal reasoning for large-scale text-based person anomaly retrieval.

\end{abstract}

\section{Introduction}
\label{sec:intro}

Text-based person retrieval (TBPS)~\cite{li2024data,yang2023aptm,zheng2020dual,nguyen2026itselfattentionguidedfinegrained} aims to retrieve a target individual from a large-scale image gallery using natural language descriptions. Existing approaches primarily rely on appearance cues, such as clothing attributes, body shape, and carried objects, to establish cross-modal correspondence between text and images. More recently, text-based person anomaly retrieval (TBAPS)~\cite{yang2025beyond} has emerged as a significantly more challenging extension of this problem, where retrieval is driven by anomalous behaviors rather than static appearance. In addition to recognizing pedestrian identity, TBAPS requires fine-grained understanding of subtle behavioral semantics, distinguishing actions with highly similar visual appearances but fundamentally different meanings, \eg, ``walking down stairs'' versus ``falling down stairs.'' This substantially increases the difficulty of cross-modal matching and demands stronger vision-language reasoning capabilities.

The AI City Challenge 2026 Track~4 formulates TBAPS on the Pedestrian Anomaly Behavior (PAB) dataset~\cite{yang2025beyond}, a large-scale benchmark consisting of over one million synthetic image--text training pairs covering 1,000 routine and 1,600 anomalous action categories. The evaluation set contains 1,978 natural-language queries and 36,773 gallery images, including 34,795 distractors, making precise retrieval particularly challenging. Performance is evaluated using Mean Average Precision (mAP), which requires both accurate ranking and robust discrimination among visually similar pedestrian instances.

Our solution builds upon the strong baseline framework of~\cite{nguyen2025hybrid}, which achieves state-of-the-art performance on the PAB benchmark through three complementary components: (1) the Local-Global Hybrid Perspective (LHP) module that enriches visual representations using probabilistic local and global image transformations; (2) Unified Image-Text (UIT) modeling, jointly optimizing image-text contrastive (ITC), image-text matching (ITM), masked language modeling (MLM), and masked image modeling (MIM) objectives; and (3) an iterative ensemble strategy that progressively refines retrieval predictions across multiple models. While highly effective, the original ensemble remains limited to a relatively homogeneous set of embedding models and does not fully exploit the complementary representations offered by recent large-scale vision-language models.

To address this limitation, we extend the baseline by incorporating diverse pre-trained vision-language embedding models, including Voyage Multimodal~\cite{voyage2025}, BGE-VL-v1.5-mmeb~\cite{zhou2024megapairs}, and Qwen3-VL-Embedding-8B~\cite{bai2025qwen3}, into the iterative ensemble framework. Since these models independently process the gallery and produce similarity matrices under different image orderings, we introduce a score alignment procedure that projects all predictions onto a unified gallery indexing, enabling valid score-level fusion. Furthermore, we propose an \textit{Ensemble Disagreement} routing mechanism that identifies ambiguous queries based on inter-model disagreement and selectively invokes a computationally expensive VLM cross-encoder reranker (gemini-3.1-flash-lite) only when necessary. The reranked predictions are subsequently integrated with the original retrieval results via Reciprocal Rank Fusion (RRF), yielding a more accurate yet computationally efficient retrieval pipeline.

Experimental results on the AI City Challenge 2026 Track~4 benchmark demonstrate the effectiveness of our approach, achieving \textbf{90.92\%} mAP, \textbf{85.13\%} Recall@1, \textbf{97.72\%} Recall@5, and \textbf{98.68\%} Recall@10.

Our contributions are summarized as follows:
\begin{itemize}
    \item We adopt the strong TBAPS baseline~\cite{nguyen2025hybrid}, leveraging its Local-Global Hybrid Perspective (LHP) and Unified Image-Text (UIT) framework for effective cross-modal representation learning.
    
    \item We enrich the iterative ensemble with complementary pre-trained vision-language embedding models, including Voyage Multimodal, BGE-VL-v1.5-mmeb, and Qwen3-VL-Embedding-8B, enabling more diverse cross-modal representations for anomaly retrieval.
    
    \item We propose a score alignment strategy that resolves heterogeneous gallery orderings across different embedding models, allowing reliable score-level fusion within the ensemble.
    
    \item We introduce an \textit{Ensemble Disagreement} routing mechanism that selectively performs VLM-based cross-encoder reranking only for uncertain queries, followed by Reciprocal Rank Fusion to produce the final retrieval results.
\end{itemize}

\section{Related Work}
\label{sec:related}

\subsection{Vision-Language Embedding Models}

Large-scale vision-language pre-training has substantially advanced cross-modal representation learning by aligning images and text within a shared embedding space. CLIP~\cite{radford2021clip} pioneered contrastive vision-language pre-training on web-scale image--text pairs, demonstrating remarkable zero-shot transferability across diverse downstream tasks. Subsequent models have further improved cross-modal understanding through stronger architectures and training paradigms. BEiT-3~\cite{wang2022beit3} formulates images as a ``foreign language,'' enabling unified multimodal pre-training with a single Transformer backbone, while BLIP-2~\cite{li2023blip2} bridges frozen vision encoders and large language models through lightweight query transformers. More recently, embedding-oriented models such as Voyage Multimodal~\cite{voyage2025} and BGE-VL~\cite{zhou2024megapairs} have demonstrated strong retrieval performance, whereas large vision-language models such as Qwen-VL~\cite{bai2025qwen3} provide enhanced multimodal reasoning capabilities. These models offer complementary semantic representations, motivating their integration within retrieval ensembles.

\subsection{Text-Based Person Retrieval and Anomaly Search}

Text-based person retrieval (TBPS) aims to retrieve target pedestrians from image galleries using natural language descriptions. Early methods, such as Dual-path CNN~\cite{zheng2020dual}, learned separate visual and textual representations and aligned them through contrastive learning. More recently, APTM~\cite{yang2023aptm} introduced large-scale multi-attribute pre-training to improve fine-grained cross-modal correspondence, significantly advancing retrieval performance.

Building upon TBPS, text-based person anomaly search (TBAPS) extends the retrieval objective from appearance matching to behavior understanding. The Pedestrian Anomaly Behavior (PAB) benchmark~\cite{yang2025beyond} established the first large-scale dataset for this task, containing over one million synthetic image--text pairs covering 1,000 routine and 1,600 anomalous behaviors. Compared with conventional TBPS datasets, PAB requires models to jointly reason about pedestrian appearance, human actions, object interactions, and surrounding scene context, making retrieval considerably more challenging. The proposed Cross-Modal Pose-aware (CMP) framework and its variants, including PE and IHNM~\cite{yang2025beyond}, serve as the first baselines for this benchmark. More recently, Nguyen \etal~\cite{nguyen2025hybrid} proposed a Hybrid, Unified, and Iterative framework that substantially improves TBAPS performance through hybrid visual augmentation, unified vision-language pre-training objectives, and iterative ensemble learning.

\subsection{Ensemble Learning for Retrieval}

Ensemble learning is a widely adopted strategy for improving retrieval robustness by exploiting the complementary strengths of multiple models. Existing approaches typically combine retrieval scores from independently trained models through weighted averaging or iterative refinement, reducing prediction variance while preserving computational efficiency~\cite{zhang2025enhancing, hu2025solution,nguyen2026escemotionalselfcorrectionreliable}. In the text-based person anomaly search (TBAPS) setting, Nguyen \etal~\cite{nguyen2025hybrid} introduced an iterative ensemble strategy that progressively refines retrieval scores across multiple models, achieving state-of-the-art performance on the PAB benchmark. Nevertheless, existing ensembles primarily aggregate models with similar architectures and embedding spaces. In contrast, our work explores heterogeneous ensembles that integrate diverse vision-language embedding models with fundamentally different representation characteristics.

\subsection{Cross-Encoder Reranking and Late Fusion}

Modern retrieval systems commonly adopt a two-stage retrieve-and-rerank pipeline to balance efficiency and accuracy~\cite{nogueira2019passage}. Lightweight dual-encoder models first retrieve a candidate set, after which more expressive cross-encoders perform fine-grained cross-modal reasoning over the top-ranked results. Recent large vision-language models (LVLMs) have demonstrated remarkable zero-shot reasoning ability, making them effective cross-encoder rerankers for complex multimodal retrieval tasks~\cite{sun2023chatgpt}. For example, AnomalyLMM~\cite{ju2025anomalylmm} bridges generative knowledge and discriminative retrieval by applying generative LVLMs for person anomaly search. However, applying generative VLM reasoning to every query is computationally expensive and slow. Consequently, our approach differs by employing an adaptive routing strategy that selectively invokes an expensive VLM reranker only for ambiguous queries identified through ensemble disagreement. This mechanism preserves dual-encoder efficiency while solving hard queries. To further combine predictions from heterogeneous retrieval models and the VLM, rank-based aggregation methods such as Reciprocal Rank Fusion (RRF)~\cite{cormack2009reciprocal} are widely adopted because they effectively fuse rankings without requiring score normalization or calibration.

\section{Method}
\label{sec:method}

\subsection{Overview}

\begin{figure*}[t]
  \centering
  \includegraphics[width=\textwidth]{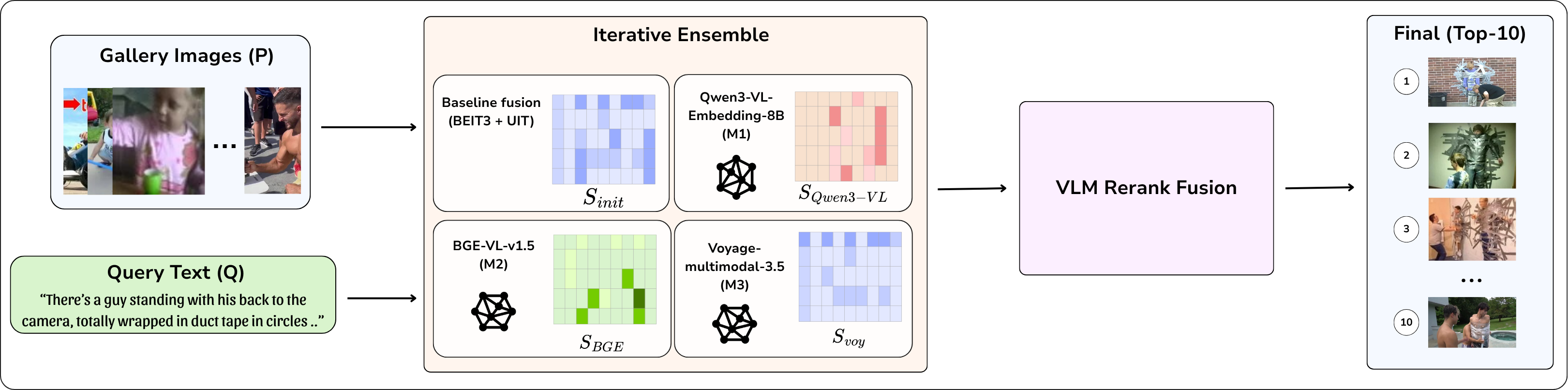}
  \caption{Overview of our \textbf{pipeline}: an iterative ensemble of embedding models refines the baseline similarity scores, and VLM rerank fusion produces the final ranking.}
  \label{fig:overview}
\end{figure*}

Our framework follows a coarse-to-fine retrieval paradigm that progressively improves retrieval quality through heterogeneous ensemble learning and selective vision-language reasoning. As illustrated in Fig.~\ref{fig:overview}, we first obtain coarse retrieval scores using a strong dual-encoder retrieval backbone. These predictions are subsequently refined by aggregating multiple heterogeneous vision-language embedding models, each providing complementary semantic representations. Since different embedding models may produce similarity matrices under different gallery orderings, we introduce a score alignment procedure that projects all similarity scores onto a unified gallery index before fusion. Finally, we employ a disagreement-aware routing mechanism that identifies ambiguous queries based on ensemble consensus and selectively invokes a large vision-language model (VLM) for cross-encoder reranking. The reranked results are then integrated with the original ensemble predictions using Reciprocal Rank Fusion (RRF), producing the final retrieval ranking.

\subsection{Base Retrieval Framework}
\label{sec:base}

We adopt the Hybrid, Unified, and Iterative framework~\cite{nguyen2025hybrid} as the foundation of our retrieval system. Specifically, the framework combines the Local-Global Hybrid Perspective (LHP) module with Unified Image-Text (UIT) learning to produce robust cross-modal representations for text-based person anomaly retrieval. Given a textual query $q$ and a gallery image $i$, the retrieval backbone encodes them into a shared embedding space,

\begin{equation}
\mathbf{z}_q=f_q(q), \qquad
\mathbf{z}_i=f_i(i),
\end{equation}

where $f_q(\cdot)$ and $f_i(\cdot)$ denote the text and image encoders, respectively. Their similarity is computed using cosine similarity,

\begin{equation}
S(i,q)=
\frac{\mathbf{z}_i^\top\mathbf{z}_q}
{\|\mathbf{z}_i\|_2\|\mathbf{z}_q\|_2}.
\end{equation}

This baseline provides a strong initial ranking that serves as the foundation for the subsequent refinement stages.

\subsection{Heterogeneous Embedding Ensemble}
\label{sec:ensemble}

% Preamble cần: \usepackage{graphicx}

\begin{figure*}[t]
  \centering
  \includegraphics[width=\textwidth]{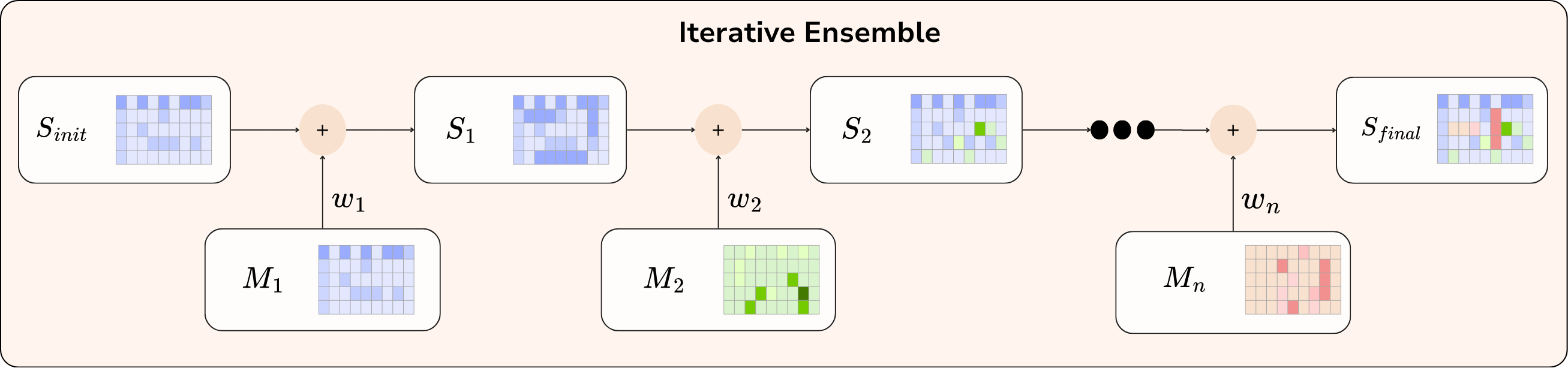}
  \caption{Overview of our \textbf{iterative ensemble}. Starting from the initial gallery embedding set $S_{\mathrm{init}}$, similarity scores are iteratively fused with those of each embedding model $M_i$ using a weighted combination, where $w_i$ is the weight assigned to $M_i$. Each step produces a refined embedding set $S_i$, and the process yields the final set $S_{\mathrm{final}}$ after $n$ steps.}
  \label{fig:iterative_ensemble}
\end{figure*}

Although modern vision-language embedding models share the common objective of aligning images and text, they are trained using different architectures, objectives, and datasets, resulting in complementary embedding spaces. Instead of relying on a single representation, we leverage multiple heterogeneous embedding models to improve retrieval robustness. An overview of our iterative ensemble is shown in Figure~\ref{fig:iterative_ensemble}.

Let $\mathcal{M}=\{M_1,M_2,\ldots,M_K\}$ denote the set of embedding models. Each model independently computes a similarity matrix
\begin{equation}
S^{(k)}\in\mathbb{R}^{N_q\times N_g},
\end{equation}
where $N_q$ and $N_g$ denote the numbers of queries and gallery images, respectively. 

Let $S_{\mathrm{init}}$ denote the initial similarity matrix obtained from the baseline retrieval model. Following the iterative ensemble formulation of~\cite{nguyen2025hybrid}, the accumulated similarity matrix $S$ is initialized as $S = S_{\mathrm{init}}$ and is progressively updated for each model $M_k \in \mathcal{M}$ as:
\begin{equation}
S \leftarrow w_k S + (1-w_k)S^{(k)},
\end{equation}
where $w_k$ controls the contribution of the accumulated predictions relative to the newly incorporated embedding model $M_k$. By iteratively integrating heterogeneous embedding spaces, the ensemble exploits complementary semantic information while preserving the strong retrieval capability of the baseline model.

\subsection{Score Alignment}
\label{sec:alignment}

Unlike models trained within a unified framework, independently developed embedding models may preprocess or index gallery images differently, resulting in inconsistent gallery orderings across similarity matrices. Consequently, directly averaging similarity scores from different models would incorrectly associate scores with mismatched gallery identities.

To enable valid score-level fusion, we introduce a score alignment procedure that projects every similarity matrix onto a shared gallery indexing. Specifically, for each embedding model, we construct a mapping between its internal gallery ordering and the reference gallery order used by the baseline retrieval system. The similarity matrix is subsequently rearranged according to this mapping before ensemble fusion. This alignment guarantees that similarity scores from different embedding models consistently correspond to the same gallery images, enabling reliable heterogeneous ensemble learning.

\subsection{Disagreement-aware VLM Reranking}
\label{sec:rerank}

While heterogeneous ensembles substantially improve retrieval robustness, dual-encoder architectures remain limited in distinguishing fine-grained anomaly semantics, particularly when multiple candidates exhibit highly similar appearance. To address this limitation, we introduce a disagreement-aware routing mechanism that selectively applies computationally expensive vision-language reasoning only when necessary.

Given the retrieval rankings produced by multiple ensemble configurations, we first measure the degree of consensus among their top predictions. Queries with high consensus are considered confident and directly retain the ensemble ranking. Conversely, queries exhibiting substantial disagreement are regarded as ambiguous and are forwarded to a large vision-language model operating as a cross-encoder reranker. Unlike dual-encoder retrieval, the cross-encoder jointly attends to both the textual query and candidate images, enabling more precise reasoning over subtle behavioral differences and complex semantic descriptions.

This adaptive routing strategy preserves the efficiency of dual-encoder retrieval while reserving expensive cross-modal reasoning for the most challenging retrieval cases.

\subsection{Final Rank Fusion}
\label{sec:rrf}

% Preamble cần: \usepackage{graphicx}

\begin{figure*}[t]
  \centering
  \includegraphics[width=\textwidth]{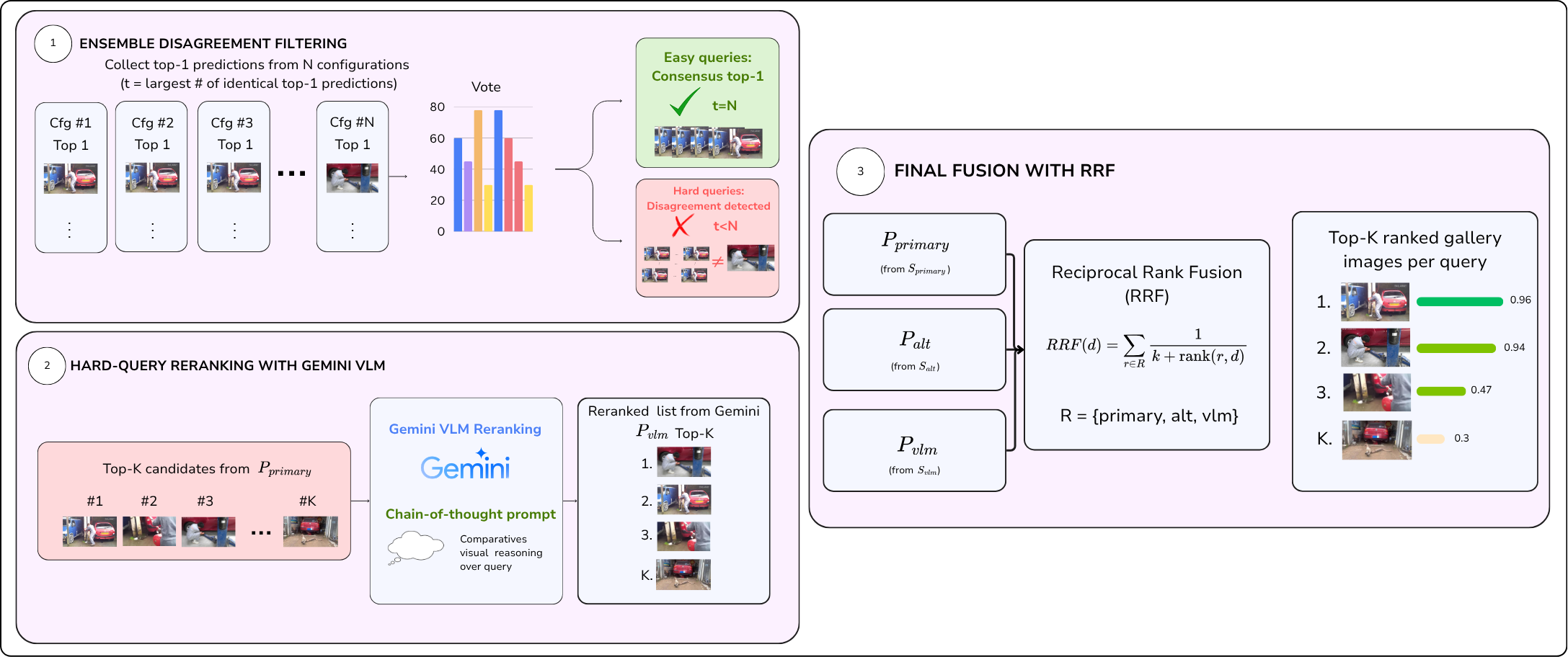}
  \caption{Our \textbf{VLM reranking and fusion pipeline}: ensemble disagreement filtering identifies hard queries, which are reranked by a VLM ($P_{VLM}$) and fused with the top-two ensemble rankings ($P_{primary}$ and $P_{alt}$) via RRF.}
  \label{fig:vlm_rerank_rrf}
\end{figure*}

The final retrieval results combine predictions from heterogeneous dual-encoder ensembles and the VLM reranker. Since these models produce similarity scores with different scales and distributions, direct score averaging is unreliable. Instead, we employ Reciprocal Rank Fusion (RRF), which aggregates ranked lists without requiring score normalization.

Given a set of ranked lists $\mathcal{R}$, the final score of gallery image $d$ is computed as

\begin{equation}
\mathrm{RRF}(d)
=
\sum_{r\in\mathcal{R}}
\frac{1}{k+\mathrm{rank}(r,d)},
\end{equation}

where $\mathrm{rank}(r,d)$ denotes the ranking position of image $d$ in ranked list $r$, and $k$ is a constant that controls the contribution of lower-ranked candidates. The final gallery ranking is obtained by sorting images according to their RRF scores as illustrated by Fig~\ref{fig:vlm_rerank_rrf}. By integrating complementary predictions from heterogeneous embedding models and selective VLM reranking, the proposed framework achieves a more robust retrieval system for text-based person anomaly search.

\section{Experiments}
\label{sec:exp}

\subsection{Dataset}
\label{sec:dataset}

We evaluate all methods on the official Pedestrian Anomaly Behavior (PAB) benchmark~\cite{yang2025beyond}, released for AI City Challenge 2026 Track~4. The benchmark is specifically designed for text-based person anomaly retrieval, where the objective is to retrieve the target pedestrian from a large gallery given a natural language description.

The training set consists of more than one million synthetic image--text pairs covering 1,000 routine behaviors and 1,600 anomalous behaviors. The evaluation set contains 1,978 textual queries and 36,773 gallery images, including 34,795 distractors collected from real-world surveillance scenarios. Compared with conventional text-based person retrieval benchmarks, the large number of visually similar distractors and the subtle distinctions between anomalous behaviors require models to jointly reason about pedestrian appearance, object interactions, and behavioral semantics.

Following the official evaluation protocol, retrieval performance is measured using Mean Average Precision (mAP), Recall@1, Recall@5, and Recall@10.

\subsection{Implementation Details}
\label{sec:impl}

Our framework is implemented using the PyTorch deep learning framework. Following the baseline of~\cite{nguyen2025hybrid}, we fine-tune the Local-Global Hybrid Perspective (LHP) module for 3 epochs and the Unified Image-Text (UIT) model for 15 epochs. The resulting retrieval backbone is subsequently used to extract image and text embeddings for the initial retrieval stage.

All training and inference are conducted on a single NVIDIA RTX A6000 GPU with 48\,GB of VRAM.

Feature extraction is performed using a combination of local GPUs, cloud computing resources, and commercial embedding APIs according to the deployment requirements of individual models. During heterogeneous ensemble learning, similarity matrices from different embedding models are first aligned to a common gallery ordering before iterative score fusion. Unless otherwise specified, all hyperparameters, including the ensemble weight schedule and reranking configuration, are determined empirically on the validation set and kept fixed throughout all experiments.

\subsection{Ablation Studies}
\label{sec:ablation}

To better understand the contribution of each component in our framework, we conduct a series of ablation studies that progressively construct the final retrieval system. Specifically, we first evaluate different retrieval backbones to determine the strongest baseline. We then investigate the effectiveness of heterogeneous vision-language embedding ensembles and finally analyze the contribution of the proposed disagreement-aware VLM reranking strategy.

\subsubsection{Baseline Framework Selection}
\label{sec:ablation_backbone}

\begin{table*}[t]
\centering
\caption{Comparison with state-of-the-art methods on the Pedestrian Anomaly Behavior (PAB) dataset.
Results are categorized by evaluation settings: Standard PAB benchmark vs. Full Challenge Test Set (which includes 34,795 distractor images, totaling 36,773 gallery images).
$\dagger$ denotes results reported from corresponding publications.
Our proposed method establishes a new state-of-the-art under the standard setting and provides a strong baseline under the highly challenging 36.7k gallery setting.}
\label{tab:comparison_pab}
\resizebox{\textwidth}{!}{%
\begin{tabular}{@{}llccccc@{}}
\toprule
\textbf{Method} & \textbf{Venue} & \textbf{Gallery Size} & \textbf{mAP (\%)} & \textbf{R@1 (\%)} & \textbf{R@5 (\%)} & \textbf{R@10 (\%)} \\
\midrule
\multicolumn{7}{@{}l}{\textit{Setting A: Standard PAB Benchmark (without large-scale distractors):}} \\
CLIP~\cite{radford2021clip}$^\dagger$ & ICML'21 & Standard & -- & 47.57 & -- & -- \\
IRRA~\cite{jiang2023cross}$^\dagger$ & CVPR'23 & Standard & -- & 78.67 & -- & -- \\
X-VLM~\cite{zeng2021multi}$^\dagger$ & ICML'22 & Standard & -- & 81.95 & -- & -- \\
RaSa~\cite{bai2023rasa}$^\dagger$ & IJCAI'23 & Standard & -- & 80.79 & -- & -- \\
CMP~\cite{yang2025beyond}$^\dagger$ & ICCV'25 & Standard & 90.41 & 84.93 & -- & -- \\
AnomalyLMM~\cite{ju2025anomalylmm}$^\dagger$ & arXiv'25 & Standard & 90.89 & 84.73 & -- & -- \\
SSDC~\cite{xie2026ssdc}$^\dagger$ & ACL'26 & Standard & 92.74 & 87.01 & -- & -- \\
HUI~\cite{nguyen2025hybrid}$^\dagger$ & WWW'25 & Standard & -- & 89.23 & 99.70 & 99.90 \\
Ours: Iterative Embedding Ensemble & -- & Standard & 94.87 & 90.09 & \textbf{99.75} & \textbf{100.0} \\
\textbf{Ours: + VLM Rerank \& RRF} & -- & Standard & \textbf{94.99} & \textbf{90.34} & \textbf{99.75} & \textbf{100.0} \\
\midrule
\multicolumn{7}{@{}l}{\textit{Setting B: Full Challenge Test Set (with 34,795 distractor images, 36,773 total):}} \\
Ours: Iterative Embedding Ensemble & -- & 36,773 & 90.90 & 85.08 & 97.72 & 98.68 \\
\textbf{Ours: + VLM Rerank \& RRF} & -- & 36,773 & \textbf{90.92} & \textbf{85.13} & \textbf{97.72} & \textbf{98.68} \\
\bottomrule
\end{tabular}%
}
\end{table*}

% \begin{table}[t]
% \centering
% \caption{Overall performance comparison on the PAB dataset under
% different training-set sizes. The best results are highlighted in bold.}
% \label{tab:main_results}
% \resizebox{\columnwidth}{!}{%
% \begin{tabular}{llccc}
% \toprule
% \textbf{Training Images}
% & \textbf{Method}
% & \textbf{R@1 (\%)}
% & \textbf{R@5 (\%)}
% & \textbf{R@10 (\%)} \\
% \midrule

% \multirow{5}{*}{0.1M}
% & APTM~\cite{yang2023aptm} (zero-shot)
% & 9.40 & 22.14 & 30.18 \\

% & APTM~\cite{yang2023aptm} (fine-tuned)
% & 69.92 & 95.60 & 97.32 \\

% & IHNM~\cite{yang2025beyond}
% & 72.25 & 95.91 & 98.03 \\

% & PE~\cite{yang2025beyond}
% & 71.79 & 95.40 & 97.83 \\

% & CMP~\cite{yang2025beyond}
% & 72.80 & 96.01 & 97.47 \\

% \midrule

% 1M
% & CMP~\cite{yang2025beyond}
% & 79.53 & 97.93 & 98.84 \\

% \midrule

% \end{tabular}%
% }
% \end{table}

To establish a strong foundation for our solution, we first review the performance of representative text-based person retrieval and person anomaly retrieval methods on the Pedestrian Anomaly Behavior (PAB) benchmark. Table~\ref{tab:comparison_pab} summarizes the reported results from the literature together with the official AI City Challenge evaluation setting.

Among existing methods, the Hybrid, Unified and Iterative framework~\cite{nguyen2025hybrid} achieves the best retrieval performance under the standard PAB evaluation, reaching 89.23\% Recall@1 while maintaining nearly perfect Recall@5 and Recall@10. Compared with earlier approaches such as CMP~\cite{yang2025beyond}, AnomalyLMM~\cite{ju2025anomalylmm}, and SSDC~\cite{xie2026ssdc}, it demonstrates superior retrieval robustness through the combination of Local-Global Hybrid Perspective (LHP), Unified Image-Text (UIT) learning, and iterative ensemble fusion. Motivated by its strong retrieval capability and modular design, we adopt this framework as the retrieval backbone and further extend it with heterogeneous vision-language embedding models and selective VLM reranking.

\subsubsection{Progressive Heterogeneous Ensemble}
\label{sec:ablation_ensemble}

% \begin{table*}[t]
% \centering
% \caption{Ablation study on candidate models for the second refinement step, with Voyage fixed as the first refinement step and Qwen fixed as the final step. All metrics are reported in \%. The best performance is achieved with BGE-VL-v1.5-mmeb as the intermediate refinement model.}
% \label{tab:ablation_ensemble}
% \renewcommand{\arraystretch}{1.15}
% \begin{tabular*}{\textwidth}{@{\extracolsep{\fill}}lccccc@{}}
% \toprule
% \textbf{Ensemble Configuration} & \textbf{Weight Ratio} & \textbf{mAP} & \textbf{R@1} & \textbf{R@5} & \textbf{R@10} \\
% \midrule
% \multicolumn{6}{@{}l}{\textit{2-step refinement} ($S_{init}$ $\rightarrow$ Qwen $\rightarrow$ Voyage):} \\
% \quad Qwen + Voyage                    & 0.9:0.92      & 88.96 & 82.15 & 96.41 & 97.97 \\
% \quad Qwen + Voyage (refined schedule) & 0.88:0.9      & 90.25 & 84.07 & 97.47 & 98.58 \\
% \midrule
% \multicolumn{6}{@{}l}{\textit{3-step refinement} ($S_{init}$ $\rightarrow$ Voyage $\rightarrow$ $M_2$ $\rightarrow$ Qwen):} \\
% \quad Qwen + Voyage + E5-Omni          & 0.9:0.92:0.88 & 89.80 & 83.31 & 97.32 & 98.53 \\
% \quad Qwen + Voyage + Rzen-VL          & 0.9:0.92:0.88 & 90.55 & 84.58 & 97.47 & 98.53 \\
% \quad \textbf{Qwen + Voyage + BGE-VL-v1.5} & 0.9:0.92:0.88 & \textbf{90.90} & \textbf{85.08} & \textbf{97.72} & 98.68 \\
% \quad Qwen + Voyage + DFN5B            & 0.9:0.92:0.88 & 90.38 & 84.22 & 97.62 & \textbf{98.73} \\
% \bottomrule
% \end{tabular*}
% \end{table*}

\begin{table*}[t]
\centering
\caption{Ablation study on candidate models for the second refinement step, with Voyage fixed as the first refinement step and Qwen fixed as the final step, evaluated on the Full Challenge Test Set (Setting B). All metrics are reported in \%. The best performance is achieved with BGE-VL-v1.5-mmeb as the intermediate refinement model.}
\label{tab:ablation_ensemble}
\renewcommand{\arraystretch}{1.15}
\begin{tabular*}{\textwidth}{@{\extracolsep{\fill}}lccccc@{}}
\toprule
\textbf{Ensemble Configuration} & \textbf{Weight Ratio} & \textbf{mAP} & \textbf{R@1} & \textbf{R@5} & \textbf{R@10} \\
\midrule
\multicolumn{6}{@{}l}{\textit{2-step refinement} ($S_{init}$ $\rightarrow$ Voyage $\rightarrow$ Qwen):} \\
\quad Voyage + Qwen                    & 0.9:0.92      & 88.96 & 82.15 & 96.41 & 97.97 \\
\quad Voyage + Qwen (refined schedule) & 0.88:0.9      & 90.25 & 84.07 & 97.47 & 98.58 \\
\midrule
\multicolumn{6}{@{}l}{\textit{3-step refinement} ($S_{init}$ $\rightarrow$ Voyage $\rightarrow$ $M_2$ $\rightarrow$ Qwen):} \\
\quad Voyage + E5-Omni + Qwen          & 0.9:0.92:0.88 & 89.80 & 83.31 & 97.32 & 98.53 \\
\quad Voyage + Rzen-VL + Qwen          & 0.9:0.92:0.88 & 90.55 & 84.58 & 97.47 & 98.53 \\
\quad \textbf{Voyage + BGE-VL-v1.5-mmeb + Qwen} & 0.9:0.92:0.88 & \textbf{90.90} & \textbf{85.08} & \textbf{97.72} & 98.68 \\
\quad Voyage + DFN5B + Qwen            & 0.9:0.92:0.88 & 90.38 & 84.22 & 97.62 & \textbf{98.73} \\
\bottomrule
\end{tabular*}
\end{table*}

Starting from the selected retrieval backbone, we progressively incorporate heterogeneous vision-language embedding models through the iterative ensemble strategy described in Section~\ref{sec:ensemble}. Table~\ref{tab:ablation_ensemble} presents the performance of different ensemble configurations.

Introducing Voyage Multimodal as the first refinement model already produces a substantial improvement over the baseline, indicating that its embedding space provides complementary semantic information beyond the original retrieval model. Furthermore, placing Qwen3-VL-Embedding as the final refinement stage yields another crucial practical insight: the ensemble is highly sensitive to the fusion weights of this large vision-language model. As shown in the 2-step refinement block of Table~\ref{tab:ablation_ensemble}, a minor adjustment to the weight schedule (from 0.9:0.92 to 0.88:0.9) results in a significant $+1.29\%$ mAP boost (from 88.96\% to 90.25\%). This demonstrates that while Qwen provides powerful cross-modal reasoning, positioning it at the very end of the pipeline with carefully tuned weights is essential to correct residual hard negatives without overriding the geometric alignment established by the preceding dual-encoders.

To maximize ensemble diversity, we evaluate several candidate embedding models as an intermediate refinement stage ($M_2$) inserted between Voyage and Qwen, including E5-Omni, Rzen-VL, BGE-VL-v1.5-mmeb, and DFN5B. As shown in Table~\ref{tab:ablation_ensemble}, incorporating BGE-VL-v1.5-mmeb yields the highest performance on the challenge setting (Setting B), demonstrating that embedding models trained with different objectives capture additional cross-modal correspondences. Interestingly, we observe that while substituting it with Rzen-VL yields even higher accuracy on the standard benchmark (Setting A, reaching 95.12\% mAP), BGE-VL-v1.5-mmeb proves more robust against the massive distractors in Setting B. This reveals a crucial practical insight: stronger standalone embedding models do not necessarily yield the best ensemble performance. Ultimately, the overall improvement largely depends on the complementarity between embedding spaces the degree to which they correct each other's blind spots rather than just individual standalone performance, motivating our final ensemble architecture for the challenge.

\subsubsection{Disagreement-aware VLM Reranking Analysis}
\label{sec:ablation_rerank}

To validate the efficacy of our disagreement-aware routing mechanism, we investigate the impact of selective VLM reranking. We adopt a zero-tolerance disagreement threshold: a query is routed to the VLM if there is any inconsistency among the top-1 predictions of the ensemble configurations. This conservative threshold is deliberately chosen to maximize recall on subtle behavioral anomalies; because dual-encoder architectures are highly susceptible to visual distractors, any prediction divergence serves as a strong indicator of an edge case requiring fine-grained cross-modal reasoning. Consequently, this strict criterion isolates 299 ambiguous queries (15.1\% of the test set). By restricting the computationally intensive \texttt{gemini-3.1-flash-lite} cross-encoder to rerank only the top-10 candidates of this challenging subset (using a prompt that instructs the model to act as an expert surveillance analyst to identify the precise anomalous behavior), our approach yields an 84.9\% reduction in inference latency compared to exhaustive reranking.

To resolve these conflicting predictions, the VLM evaluates the top-10 candidates for each routed query, rectifying prediction inconsistencies and validating ensemble consensus. The final ranking is aggregated using Reciprocal Rank Fusion (RRF) with $k{=}60$. While the absolute mAP gain is modest, it is crucial to recognize that the routed queries encapsulate the most pathological edge cases instances where multiple robust feature representations fundamentally disagree. The capacity of the VLM to disentangle such ambiguity, while preserving the high-confidence dual-encoder predictions for the remaining 84.9\% of the dataset (1,679 out of 1,978 queries), underscores the representational synergy and computational efficiency of our adaptive routing strategy.

\subsection{Final Challenge Results}
\label{sec:results}

\begin{table}[t]
\centering
\caption{Performance of our final submission on the official AI City Challenge 2026 Track~4 benchmark. ($^{*}$)~denotes the model is used without fine-tuning (zero-shot). Best results are in \textbf{bold}.}
\label{tab:final_results}
\setlength{\tabcolsep}{6pt}
  \begin{tabular}{@{}lcccc@{}}
    \toprule
    \textbf{Method} & \textbf{mAP} & \textbf{R@1} & \textbf{R@5} & \textbf{R@10} \\
    \midrule
    UIT$^{*}$ + LHP$^{*}$              & 72.75 & 61.22 & 87.76 & 92.61 \\
    UIT$^{*}$ + LHP                    & 85.17 & 76.89 & 95.39 & 97.11 \\
    UIT + LHP                          & 87.24 & 80.03 & 96.00 & 97.11 \\
    UIT + LHP + BLIP-2 + CLIP          & 83.02 & 73.60 & 94.69 & 97.01 \\
    \midrule
    Ours: Iterative Ensemble           & 90.90 & 85.08 & 97.72 & 98.68 \\
    \textbf{Ours: Iterative Ensemble + Rerank Fusion}         & \textbf{90.92} & \textbf{85.13} & \textbf{97.72} & \textbf{98.68} \\
    \bottomrule
  \end{tabular}
\end{table}

Table~\ref{tab:final_results} reports the performance of our final submission on the official AI City Challenge 2026 Track~4 benchmark. Starting from the selected Hybrid retrieval backbone, we progressively improve retrieval performance through heterogeneous vision-language embedding ensemble learning. The proposed disagreement-aware VLM reranking and Reciprocal Rank Fusion further refine the predictions for ambiguous queries, leading to additional performance gains.

Our final system achieves \textbf{90.92\%} mAP, \textbf{85.13\%} Recall@1, \textbf{97.72\%} Recall@5, and \textbf{98.68\%} Recall@10. These results demonstrate that combining complementary embedding spaces with selective vision-language reasoning provides an effective solution for large-scale text-based person anomaly retrieval under challenging surveillance scenarios.

\section{Discussion}
\label{sec:discussion}

Our results demonstrate that the effectiveness of the proposed framework stems from the complementary nature of heterogeneous vision-language representations. Rather than relying on a single embedding model, progressively integrating models trained with different objectives and data distributions consistently improves retrieval performance, suggesting that representational diversity is more valuable than simply selecting the strongest individual model. This observation is further supported by our ablation study, where the best-performing ensemble is not necessarily composed of the highest-performing standalone models.

Another practical insight is the importance of score alignment for heterogeneous ensemble learning. Since independently developed embedding models may adopt different gallery orderings, direct score-level fusion can produce incorrect correspondences. Aligning all similarity matrices to a unified gallery indexing enables reliable integration of heterogeneous embedding models while remaining agnostic to the underlying architectures.

The PAB benchmark also highlights the intrinsic difficulty of text-based person anomaly retrieval. Compared with conventional text-based person retrieval, successful retrieval requires jointly reasoning about pedestrian appearance, fine-grained behaviors, object interactions, and scene context under substantial domain shifts between synthetic training data and real-world surveillance images.

\paragraph{Limitations.}
Our framework increases inference cost by combining multiple embedding models and selective VLM reranking. Moreover, the ensemble weights and routing strategy are empirically determined. Future work may explore adaptive ensemble weighting and learned query routing to further improve both retrieval accuracy and computational efficiency.

\section{Conclusion}
\label{sec:conclusion}

In this paper, we presented our solution for AI City Challenge 2026 Track~4 on text-based person anomaly retrieval. Building upon a strong retrieval backbone, we proposed a heterogeneous ensemble framework that integrates complementary vision-language embedding models through score alignment and iterative fusion, followed by selective VLM-based reranking for ambiguous queries.

Our final system achieved \textbf{90.92\%} mAP, \textbf{85.13\%} Recall@1, \textbf{97.72\%} Recall@5, and \textbf{98.68\%} Recall@10 on the official PAB benchmark. These results demonstrate the effectiveness of combining diverse vision-language representations with adaptive multimodal reasoning for fine-grained anomaly retrieval. We believe the proposed framework provides a practical and extensible solution for large-scale text-based person retrieval and offers useful insights for future multimodal retrieval systems in real-world surveillance scenarios.

\section*{Acknowledgements}
We thank the organizers of the AI City Challenge 2026 for providing the PAB dataset and evaluation platform. We acknowledge the authors of the baseline framework~\cite{nguyen2025hybrid} for making their code publicly available. We are also grateful to AIVN and GenAI4E for their valuable support, guidance, and resources throughout this work.

\bibliographystyle{splncs04}
\bibliography{main}
\end{document}